\documentclass[10pt,journal,compsoc]{IEEEtran}
\ifCLASSOPTIONcompsoc
 \usepackage{cite}
\else
 \usepackage{cite}
\fi
\usepackage{xspace}
\usepackage{tabularx, booktabs}

\ifCLASSINFOpdf
 \usepackage[pdftex]{graphicx}
 \graphicspath{{Images/}}
 \DeclareGraphicsExtensions{.pdf,.jpeg,.png}
\else
\fi
\usepackage{booktabs}

\usepackage{algorithmic}

\usepackage{xcolor}

\ifCLASSOPTIONcompsoc
 \usepackage[caption=false,font=footnotesize,labelfont=sf,textfont=sf]{subfig}
 
\else
 \usepackage[caption=false,font=footnotesize]{subfig}
\fi
\usepackage{caption}
\usepackage{color}
\usepackage[algo2e,ruled,vlined]{algorithm2e}
\usepackage{amsmath}
\usepackage{amssymb}
\usepackage{hyperref}
\usepackage{xcolor}
\usepackage[]{algorithm}
\usepackage{comment}
\usepackage{multirow}

\newcommand*{\colorboxed}{}
\def\colorboxed#1#{%
 \colorboxedAux{#1}%
}
\newcommand*{\colorboxedAux}[3]{%
 \begingroup
 \colorlet{cb@saved}{.}%
 \color#1{#2}%
 \boxed{%
 \color{cb@saved}%
 #3%
 }%
 \endgroup
}

\usepackage{array}
\newcolumntype{L}[1]{>{\raggedright\let\newline\\\arraybackslash\hspace{0pt}}m{#1}}
\newcolumntype{C}[1]{>{\centering\let\newline\\\arraybackslash\hspace{0pt}}m{#1}}
\newcolumntype{R}[1]{>{\raggedleft\let\newline\\\arraybackslash\hspace{0pt}}m{#1}}

\newcommand{\beq}{\begin{equation}}
\newcommand{\eeq}{\end{equation}}

\newcommand{\mr}[1]{\mathrm{#1}}

\begin{document}
%

\title{Multi-region segmentation of bladder cancer structures in MRI with progressive dilated convolutional networks}
%
%
%
%

\author{Jose~Dolz$^{*,\ddag}$,
Xiaopan~Xu$^*$,
J\'er\^{o}me~Rony,
Jing~Yuan,
Yang~Liu,
Eric~Granger,
Christian~Desrosiers,
Ismail~Ben~Ayed$^{\ddag}$,
 and~Hongbing~Lu$^{\ddag}$
\IEEEcompsocitemizethanks{\IEEEcompsocthanksitem J. Dolz, J. Rony, E. Granger, C. Desrosiers and I. Ben Ayed are with the ETS Montreal, Canada. X. Xu, Y. Liu, X. Zhang and H. Lu are with the School of Biomedical Engineering, Fourth Military Medical University, Xi'an, China. J. Yuan is with the School of Mathematics and Statistics, Xidian Univeristy, China.}

\thanks{$^*$J. Dolz and X. Xu contributed equally. \newline $^{\ddag}$J. Dolz, I. Ben Ayed and H. Lu are corresponding authors. Email: jose.dolz@etsmtl.ca, ismail.benayed@etsmtl.ca, luhb@fmmu.edu.cn}}


\IEEEtitleabstractindextext{%
\begin{abstract}

Precise segmentation of bladder walls and tumor regions is an essential step towards non-invasive identification of tumor stage and grade, which is critical for treatment decision and prognosis of patients with bladder cancer (BC). However, the automatic delineation of bladder walls and tumor in magnetic resonance images (MRI) is a challenging task, due to important bladder shape variations, strong intensity inhomogeneity in urine and very high variability across population, particularly on tumors appearance. To tackle these issues, we propose to leverage the representation capacity of deep fully convolutional neural networks.  The proposed network includes dilated convolutions to increase the receptive field without incurring extra cost nor degrading its performance. Furthermore, we introduce progressive dilations in each convolutional block, thereby enabling extensive receptive fields without the need for large dilation rates. The proposed network is evaluated on 3.0T T2-weighted MRI scans from 60 pathologically confirmed patients with BC.  Experiments show the proposed model to achieve a higher level of accuracy than state-of-the-art methods, with a mean Dice similarity coefficient of 0.98, 0.84 and 0.69 for inner wall, outer wall and tumor region segmentation, respectively. These results represent a strong agreement with reference contours and an increase in performance compared to existing methods. In addition, inference times are less than a second for a whole 3D volume, which is between 2-3 orders of magnitude faster than related state-of-the-art methods for this application. We showed that a CNN can yield precise segmentation of bladder walls and tumors in BC patients on MRI. The whole segmentation process is fully-automatic and yields results similar to the reference standard, demonstrating the viability of deep learning models for the automatic multi-region segmentation of bladder cancer MRI images. 

\end{abstract}

}

\maketitle

\IEEEdisplaynontitleabstractindextext

\IEEEpeerreviewmaketitle

\IEEEraisesectionheading{\section{Introduction}\label{sec:introduction}}

Urinary bladder cancer (BC) is a life-threatening disease with high morbidity and mortality rate \cite{Antoni2017Bladder,Woo2017Diagnostic,Alfred2016Updated,AmericanCancerSociety2016,Kamat2016,Choi2015}. Accurate identification of tumor stage and grade is of extreme clinical importance for the treatment decision and prognosis of patients with BC \cite{Kamat2016,Choi2015,Knowles2015,Cancer2014,Duan2010}. Clinical standard reference for this task is optical cystoscopy (OCy) with transurethral resection (TUR) biopsies, however, this procedure is often limited due to its invasiveness and discomfort for patients. With narrow field-of-view (FOV) for lumen observation and local characterization of tissue samples, single transurethral biopsy has exhibited relatively high misdiagnosis rates, especially for staging \cite{Kamat2016,Knowles2015,Cancer2014,Duan2010}. Recent advances in magnetic resonance imaging (MRI) and image processing technologies have made radiomics methods that predict tumor stage and grade using image features a potential alternative for non-invasive evaluation of BC \cite{Duan2010,qin2014adaptive,Xu2017,Xu2017Preoperative,Zhang2017Radiomics}.

Previous studies indicate that radiomics descriptors from inner and outer bladder wall (IW and OW) as well as attached tumor regions in MRI images have great potential in reflecting tumorous subtypes, properties or muscle invasiveness \cite{Xu2017Preoperative,Zhang2017Radiomics,xiao20163d,Xu2017,qin2014adaptive,duan2012adaptive}. The segmentation of bladder walls and tumors is an important step toward extracting these useful descriptors 
\cite{Duan2010,qin2014adaptive,Xu2017Preoperative,Lambin2017Radiomics}. Nowadays, the standard reference in clinical routine to segment BC structures in MRI images is based on manual delineation by experts, which is performed in a slice-by-slice manner \cite{Duan2010,qin2014adaptive,Xu2017Preoperative}. However, it is a tedious process requiring a huge amount of human effort. Furthermore, the expert's experience, imaging parameters like slice thickness, as well as image noise, motion artifacts, weak wall boundaries, or muscle-invasive tumors can impact the delineation efficiency and consistency, which may affect the radiomics-based prediction of BC properties \cite{Duan2010,qin2014adaptive,xu2017simultaneous,Xu2017Preoperative,Lambin2017Radiomics}. Therefore, an accurate and automatic multi-region segmentation tool is highly desired for tumor prediction and prognosis in a radiomics perspective \cite{Duan2010,qin2014adaptive,xu2017simultaneous,Xu2017Preoperative,Zhang2017Radiomics,Lambin2017Radiomics}.

\begin{figure}[t!]
\begin{center}
\includegraphics[width=0.85\linewidth]{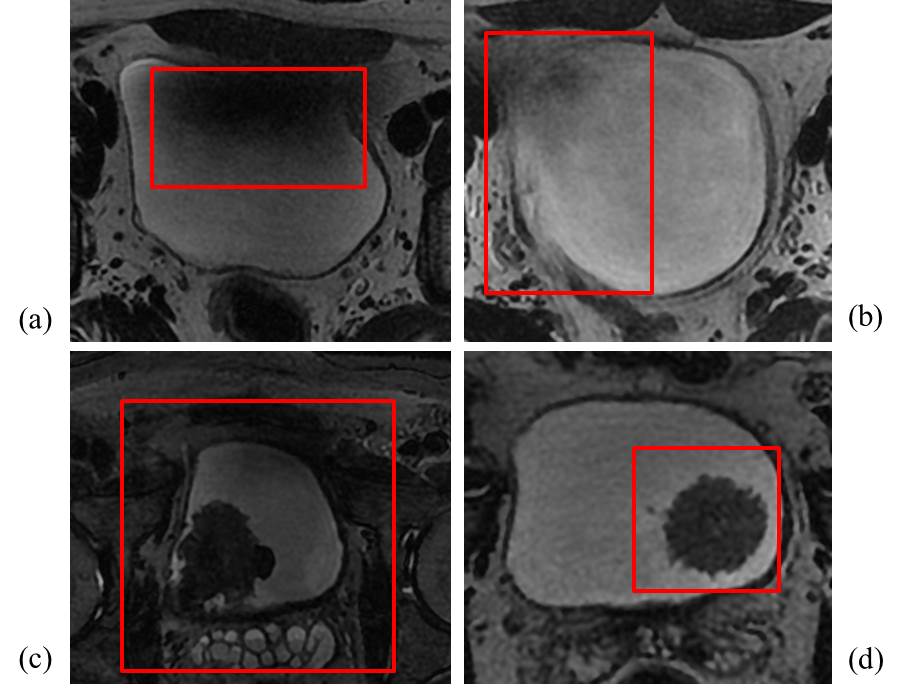}
\caption{Challenges in computer-assisted segmentation of bladder MR images (highlighted in red boxes). Image regions in red boxes represent: (a) intensity inhomogeneity in the lumen, (b) weak wall boundaries, (c) complex background intensity distribution, and (d) disconnected tumor region in the lumen, respectively.}
\label{fig:Fig1_chanllenge}
\end{center} 
\end{figure}

\begin{table*}[t!]
\centering
\footnotesize
\caption{Survey of methods to segment regions of interest in BC images via MRI.}
\label{table:methods}
\setlength{\tabcolsep}{10pt}
\begin{tabular}{llcc}
\toprule
   \textbf{Work} & \textbf{Method} & \textbf{Target}  \\
   \midrule \midrule
Li et al.,2004 \cite{li2004new}  &  Markov random field (MRF)    &  IW \\
Li et al., 2008 \cite{li2008segmentation}  &   Markov random field (MRF)      &    IW     \\
Duan et al., 2010 \cite{Duan2010}  &   Coupled Level-sets     &   IW/OW     \\
Chi et al., 2011 \cite{chi2011segmentation}  &  Coupled Level-sets       &   IW/OW        \\
Garnier et al., 2011 \cite{garnier2011bladder}  &  Active region growing  & IW   \\
Ma et al., 2011 \cite{ma2011novel}  & Geodesic active contour (GAC) + Shape-guided Chan-Vese   &   IW/OW          \\
Duan et al., 2012 \cite{duan2012adaptive}  & Coupled Level-set + Bladder wall thickness Prior     &  Tumour      \\ 
Han et al., 2013 \cite{han2013unified}  &   Adaptive Markov random field (MRF) + Coupled Level-set    &   IW/OW        \\
Qin et al., 2014 \cite{qin2014adaptive}  &   Coupled directional Level-sets     &    IW/OW   \\
Xiao et al., 2016 \cite{xiao20163d}  &   Coupled directional Level-sets + Fuzzy c-means    &   IW/OW/Tumour            \\
Xu et al., 2017 \cite{xu2017simultaneous}  &   Continuous max-flow + Bladder wall thickness Prior   &  IW/OW              \\
\bottomrule
\end{tabular}

\end{table*}

The automatic delineation of IW and OW in MRI images remains a challenging task due to important bladder shape variations, strong intensity inhomogeneity in urine caused by motion artifacts, weak boundaries and complex background intensity distribution (Fig. \ref{fig:Fig1_chanllenge}) \cite{Duan2010,qin2014adaptive,duan2012adaptive}. When further considering the presence of cancer, the problem becomes much harder as it introduces more variability across population. That might explain why literature on multi-region bladder segmentation remains scarce, with few techniques proposed to date (Table \ref{table:methods}). Initial attempts considered the use of Markov Random Fields to tackle the segmentation of the IW \cite{li2004new,li2008segmentation}. Garnier et al. \cite{garnier2011bladder} proposed a fast deformable model based on active region growing that solved the leakage issue of standard region growing algorithms. The algorithm combined an inflation force, which acts like a region growing process, and an internal force that constrains the shape of the surface. However, it would be difficult to apply these approaches directly to OW segmentation due to the complex distribution of tissues surrounding the bladder. 

Several level-set based segmentation methods have also been introduced to extract both inner and outer bladder walls \cite{Duan2010,chi2011segmentation,han2013unified,qin2014adaptive}. In \cite{Duan2010}, Duan et al. developed a coupled level-set framework which adopts a modified Chan-Vese model to locate both IW and OW from T1-weighted MRI in a 2D slice fashion. Recently, Chi et al. \cite{chi2011segmentation} applied a geodesic active contour (GAC) model in T2-weighted MRI images to segment the IW, and then coupled the constraint of maximum wall thickness in T1-weighted MRI images to segment the OW. The limitation of this work arises from the difficulty to register slices between the two sequences. To overcome these limitations, Qin et al. \cite{Qin2014} proposed an adaptive shape prior constrained level-set algorithm that evolves both IW and OW simultaneously from T2-weighted images. Despite its precision, this algorithm can be sensitive to the initializing process. In an extension of these approaches, Xiao et al. \cite{xiao20163d} introduced a second step based on fuzzy c-means \cite{bezdek1984fcm} to include tumor segmentation in the pipeline. However, this extended method showed inconsistent results between different datasets. While popular, level-sets present some important drawbacks. First, these variational approaches are based on local optimization techniques, making them highly sensitive to initialization and image quality. Second, if multiple objects are embedded in another object, multiple initializations of the active contours are required, which is time-consuming. Third, if there exist some gaps in the target, evolving contours may leak into those gaps and represent objects with incomplete contours. Finally, processing times can be prohibitive, particularly in medical applications where segmentation is typically performed in volumes. As reported in previous works, segmentation times usually exceed 20 minutes for a single 3D volume. As alternative, a modified Geodesic active contour (GAC) model and a shape-guided Chan-Vese model were proposed in \cite{ma2011novel} to segment bladder walls. Recently, Xu et al. \cite{xu2017simultaneous} introduced a continuous max-flow framework with global convex optimization to achieve a more accurate segmentation of both IW and OW. Nevertheless, the high sensitivity to initialization of all previous methods makes the full automation of segmentation very challenging. Further, most methods focus only on bladder walls and are unable to segment simultaneously both bladder walls and tumors.

Deep learning has recently emerged as a powerful modeling technique, demonstrating significant improvements in various computer vision tasks such as image classification \cite{huang2017densely}, object detection \cite{redmon2016yolo9000} and semantic segmentation \cite{yu2015multi}. Particularly, convolutional neural networks (CNNs) have been applied with enormous success to many medical image segmentation problems \cite{DolzNeuro2017,Fechter_Esophagus,litjens2017survey,dolz2018hyperdense,carass2018comparing}. Bladder segmentation was also addressed with deep learning techniques, however, the image modality of study has mainly been limited to computed tomography (CT) \cite{cha2016urinary,cha2016bladder,men2017automatic}. For example, Cha et al. \cite{cha2016urinary} proposed a CNN followed by a level-set method to segment the IW and OW. Considering the significant advantages of MRI, including its high soft-tissue contrast and non-radiation, it may be more suitable for the characterization of bladder wall and tumor properties. Surprisingly, the application of deep learning to the multi-region segmentation of bladder cancer in MRI images remains, to the best of our knowledge, unexplored.

In light of limitations with state-of-the-art methods, and inspired by the success of deep learning in medical image segmentation, we propose to address the task of multi-region bladder segmentation in MRI using a CNN. Specifically, we use a deep CNN that builds on UNet \cite{ronneberger2015u}, a well establish model for segmentation which combines a contracting path and an expansive path to get a high-resolution output of the same size as the input. To increase the receptive field spanned by the network, we propose to use a sequence of progressive dilation convolutional layers. As motivated in \cite{hamaguchi2018effective}, aggressively increasing dilation factors might fail to aggregate local features due to sparsity of the kernel, which is detrimental for small objects. Thus, we hypothesize that slowly increasing the dilation rate along the convolutions within each block may decrease sparsity in the dilated kernels, thereby allowing to capture more context while preserving the resolution of the analyzed region. As both large and small tumor regions are present in the images of the current study, this scenario benefits to both cases. This strategy enables us to span broader regions of input images without incorporating large dilation rates that can degrade segmentation performance. The current work is the first attempt to apply CNNs for multi-region segmentation of bladder cancer in MRI.

\section{Methods}\label{sec:methods}

\subsection{Fully convolutional neural networks}

CNNs are a special type of artificial neural networks that learn a hierarchy of increasingly complex features by successive convolution, pooling and non-linear activation operations \cite{krizhevsky2012imagenet,lecun1998gradient}. Originally designed for image recognition and classification, CNNs are now commonly used in semantic image segmentation. A naive approach follows a sliding-window strategy where regions defined by the window are processed independently. This technique presents two main drawbacks: reduction of segmentation accuracy and low efficiency. An alternative approach, known as fully CNNs (FCNs) \cite{FCN}, mitigates these limitations by considering the network as a single non-linear convolution that is trained in an end-to-end fashion. An important advantage of FCNs compared to standard CNNs is that they can be applied to images of arbitrary size. Moreover, because the spatial map of class scores is obtained in a single dense inference step, FCNs can avoid redundant convolution operations, making them computationally more efficient. 

The networks explored in this work are built on the UNet architecture, which has shown outstanding performance in various medical segmentation tasks \cite{christ2016automatic,cciccek20163d,dong2017automatic,sirinukunwattana2017gland,zotti2017gridnet}. This network consists of a contracting and expanding path, the former collapsing an image down into a set of high level features and the latter using these features to construct a pixel-wise segmentation mask. The original architecture also proposed skip-connections between layers at the same level in both paths, by-passing information from early feature maps to the deeper layers in the network. These skip-connections allow incorporating high level features and fine pixel-wise details simultaneously. 

Unlike in natural images, targeting clinical structures in segmentation tasks requires a certain knowledge of the global context. This information may, for example, indicate how an organ is arranged with respect to other ones. Standard convolutions have difficulty integrating global context, even when pooling operations are sequentially added into the network. For instance, in the original UNet model, the receptive field spanned by the deepest layer is only 128$\times$128 pixels. This means that the context of the entire image is not fully considered in the deep architecture to generate its final prediction. A straightforward solution for increasing the receptive field is to include additional pooling operations in the network. However, this strategy usually decreases the performance since relevant information is lost in the added down-sampling operations.

\subsection{Dilated convolutions}

\begin{figure}[ht!]
\centering
\includegraphics[width=1\linewidth]{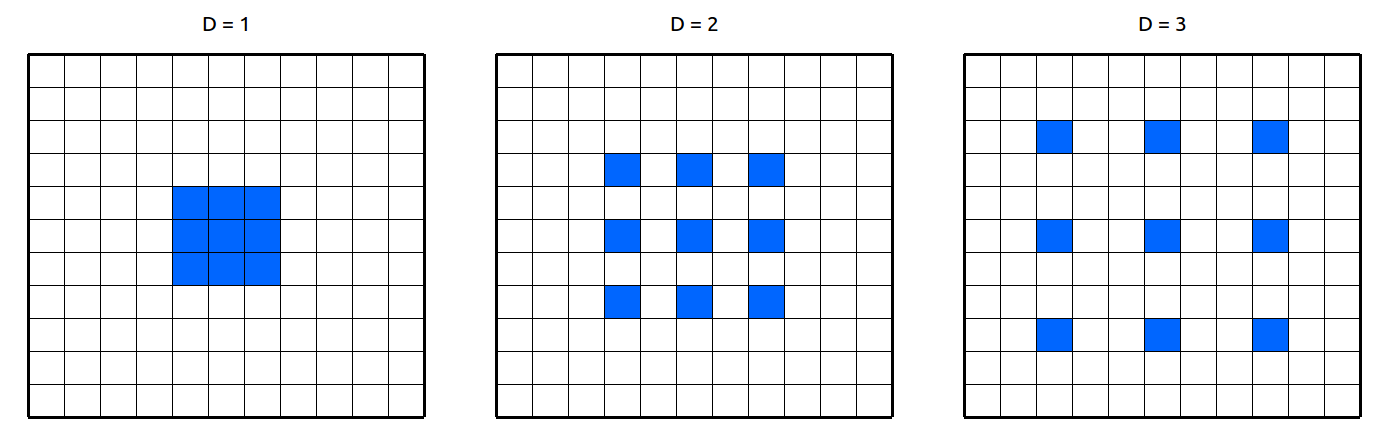}
\caption{Examples of dilation kernels and their effect on the receptive field. It is important to note that the number of parameters associated with each layer is identical.}
\label{fig:Dilations}
\end{figure}

The dilated convolution operator, also referred in the literature as \emph{a trous} convolution, was originally developed for wavelet decomposition \cite{holschneider1990real}. Very recently, Yu and Koltun \cite{yu2015multi} adopted this operation for semantic segmentation to increase the receptive field of deep CNNs, as alternative to down-sampling feature maps. The main idea is to insert ``holes'' (i.e., zeros) between pixels in convolutional kernels to increase image resolution of intermediate feature maps, thus enabling dense feature extraction in deep CNNs with an enlarged field of convolutional kernels (Fig.~\ref{fig:Dilations}). This ultimately leads to more accurate predictions \cite{wolterink2016dilated,wu2016high,moeskops2017adversarial,lopez2017dilated,chen2018deeplab,anthimopoulos2018semantic}. 

Consider a convolutional kernel $K^l$ in layer $l$ with a size of $k\times k$. The receptive field of $K^l$, also known as effective kernel size, can be defined as
\begin{equation}
 \mr{RF}_{K_l} = k + (k-1)\times(D_k-1)
\end{equation}
where $D_k$ represents the dilation rate of kernel $K_l$, specifying the number of zeros (or holes) to be placed between pixels. Note that, in standard convolutions, $D_k$ is equal to 1. Furthermore, the stride is considered equal to 1 for simplicity.

\subsection{Architecture details}

In this study, we propose to use dilated convolutions in a CNN architecture based on UNet. To evaluate the impact of dilated convolutions on segmentation performance, several models are investigated. First, we employ the original UNet implementation and a modified version that will serve as baselines (Section \ref{sssec:Unet}). Then, in the third network, the first standard convolution of each block in the baseline model is replaced by a dilated convolution (Section \ref{sssec:Unet_Dilated}). For the proposed model, the entire standard block in the baseline is replaced by the proposed progressive dilated convolutional block (Section \ref{sssec:Unet_Dilated_Prog}). Furthermore, we compare the proposed method to state-of-the-art segmentation architectures in the literature. Particularly, we investigated related deep convolutional neural networks that include factorized convolutions, including ERFNet \cite{romera2017efficient} and ENet \cite{paszke2016enet}.

\subsubsection{UNet baselines}
\label{sssec:Unet}

\begin{figure}[ht!]
\centering
\includegraphics[width=1.0\linewidth,trim={5mm 0 35mm 0},clip]{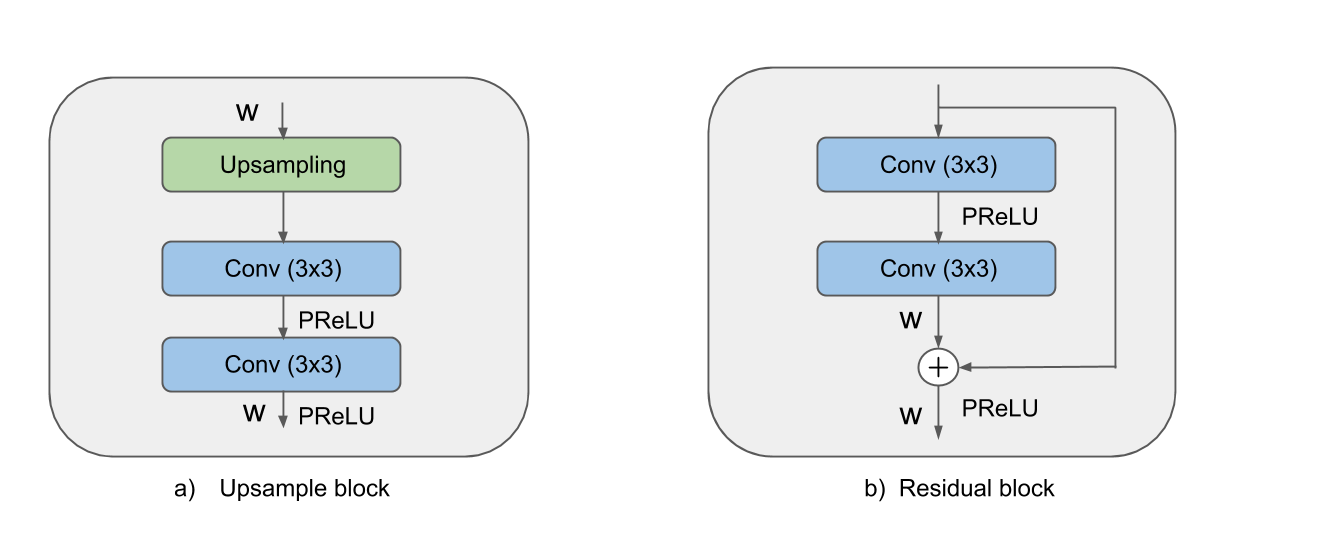}
\caption{Diagram depicting some of the blocks employed in our architectures. $W$ denotes the width or number of feature maps.}
\label{fig:detailsCNN}
\end{figure}

We employ the original version of UNet as described in \cite{ronneberger2015u} -- with the exception of using 32 kernels in the first layer instead of 64 -- as baseline network, which will be denoted as \emph{UNet-Original}. In addition, we further included three main modifications on the original version in a second baseline, denoted as \emph{UNet-Baseline}. First, we employ convolutions with stride 2 instead of max-pooling in the contracting path. Second, the deconvolutional blocks in the decoding path are replaced by upsampling and convolutional blocks, which have demonstrated to improve performance \cite{badrinarayanan2017segnet} (Fig. \ref{fig:detailsCNN},a). Third, to have a more compact representation of learned features, a bottleneck block with residual connections (Fig. \ref{fig:detailsCNN},b) is introduced between the contracting and expanding paths. The objective of these connections is to have the information flow from the block's input to its output without modification, thus encouraging the path through non-linearities to learn a residual representation of the input data \cite{he2016deep}. In addition, the number of kernels in the first convolutional block has been reduced from 64 to 32, since no improvement was observed with the heavier model, and allowing us to obtain a more efficient model.

Furthermore, each convolution layer in the proposed models performs batch normalization \cite{ioffe2015batch}. By reducing variations between training samples in mini-batch learning, this technique was shown to accelerate convergence of the parameter learning process and make the model more robust in testing. In addition, all activation functions in our networks are parametric rectifier linear units (PReLUs) \cite{he2015delving}.

\subsubsection{Dilated UNet}
\label{sssec:Unet_Dilated}

\begin{figure}[ht!]
\centering
\includegraphics[width=0.85\linewidth]{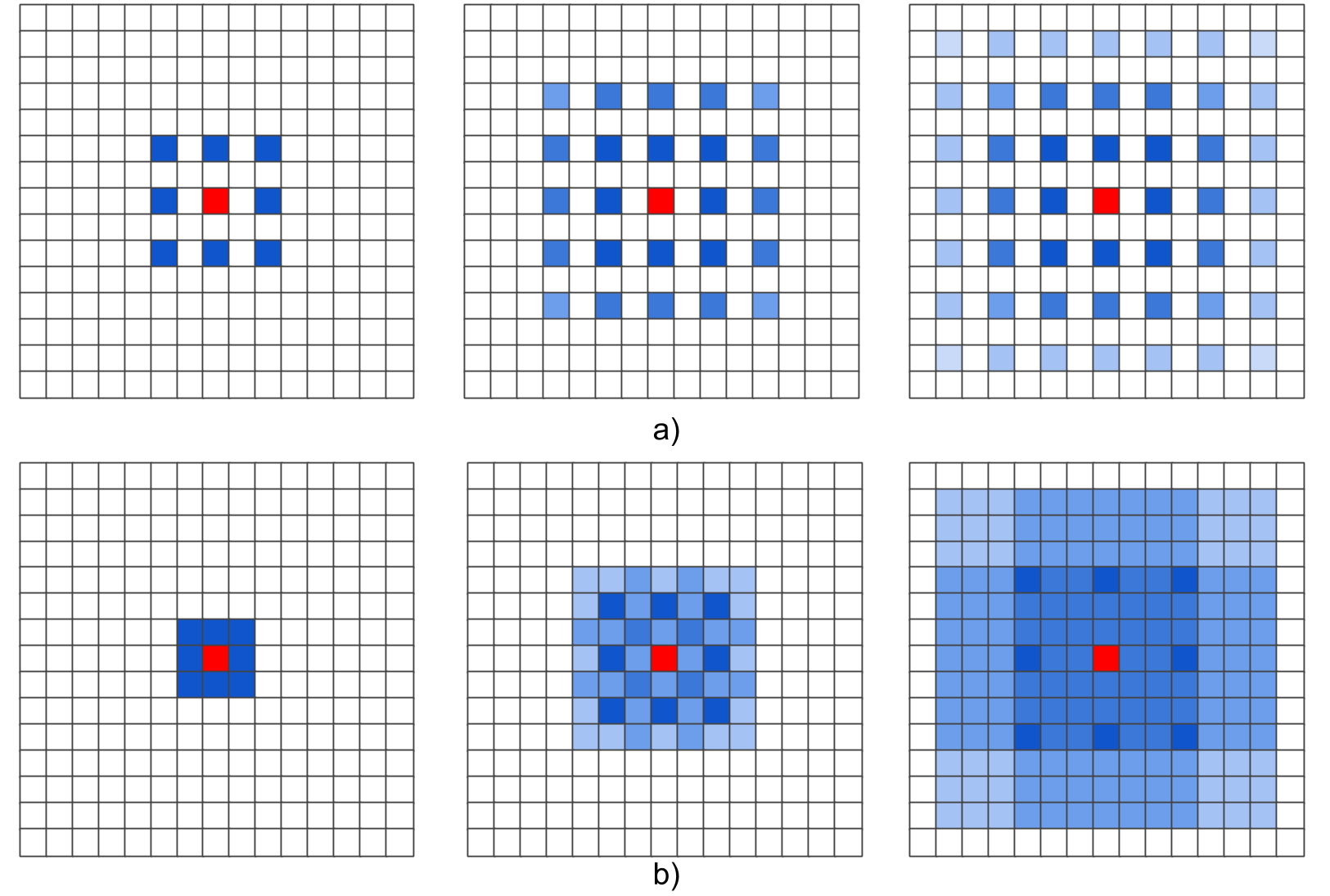}
\caption{Examples of dilation kernels and their effect on the receptive field (in blue), which illustrates the \emph{gridding} problem (\emph{a}). Each row depicts a sequence of three convolutional layers with kernel size 3$\times$3. Blue pixels contribute to the calculation of the center pixel, in red. If the dilation rate remains the same through the three convolutional layers ($r$=2), there is a loss of neighbor information due to zeros being padded between pixels. If dilation rates are exponentially increased instead ($r$=1,2,4) (\textit{b}) this effect is alleviated. Shades of blue represent the overlapping on the receptive fields of each kernel point, in dark blue. It is important to note that the number of parameters associated with each layer is identical. Image from \cite{yu2015multi}.}
\label{fig:Gridding}
\end{figure}

Our first dilated CNN model follows the general architecture of UNet, but introduces a context module at each block of the encoding path. The context module contains a dilated convolution as first operation of each block to systematically aggregate multi-scale contextual information. An inherent problem when employing dilated convolutions is \emph{gridding} \cite{wang2017understanding} (Fig. \ref{fig:Gridding},\emph{top}). As zeros are padded between pixels in a dilated convolutional kernel, the receptive field spanned by this kernel only covers an area with some sort of checkerboard patterns, sampling only locations with non-zero values. This results in the loss of neighboring information, which might be relevant for an effective learning. If dilation rate $D$ increases, this issue becomes even worse, as the convolution kernel becomes too sparse to capture any local information. To alleviate this problem, we follow the strategy proposed in other works \cite{paszke2016enet,romera2017efficient}, where dilated convolutions are alternated with standard convolutions and dilation rates are progressively increased. Therefore, the dilation rate $D$ in the convolutional blocks of this model are equal to 1, 2, 4 and 8, from shallow to deep layers, respectively.

\subsubsection{UNet with progressive dilated convolutional blocks}
\label{sssec:Unet_Dilated_Prog}

Instead of gradually increasing the dilation factor $D$ through different layers, we propose to increase it within each context module. The main idea is that features learned at each block are able to capture multi-scale level information. Therefore, at each block, the dilation rate $D$ will be equal to 1,2, and 4. With this, we avoid including large $D$ values that span broader regions, while maintaining the same network receptive field.

\begin{figure*}[ht!]
\centering
\includegraphics[width=1\linewidth]{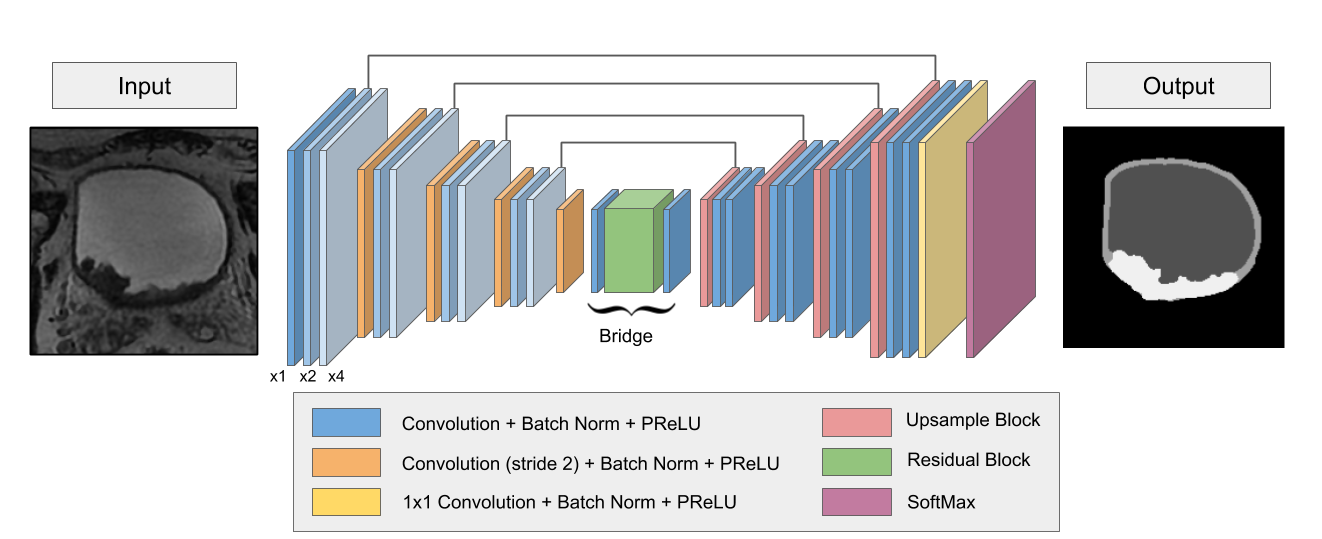}
\caption{Overall framework of the proposed deep model. Convolutional blocks on the encoding path contain progressive dilated convolutions. At each convolutional block, the dilated convolutions with ratios $\times$1, $\times$2 and $\times$4 are included. Except the last convolutional layer, in yellow, the rest of layers contain 3$\times$3 filters.}
\label{fig:Net_Gen}
\end{figure*}

Figure \ref{fig:Net_Gen} gives the schematic of the proposed model. As shown, the deep network consists of two important components: an encoder and a decoder part. While the encoder path learns the visual features for the input data, the decoder path is responsible for creating the dense segmented mask and recovering the original resolution. The encoder path is composed of 4 convolutional blocks, each one containing 3 convolutional layers, followed by a strided convolution and a bridge block which contains two convolutional layers and a residual block (Fig. \ref{fig:detailsCNN},b) -- generating a structure with a depth of 16 layers. On the other hand, the decoding path has 17 convolutional layers distributed as follows: 4 upsampling modules, which contains 4 convolutional layers each, followed by a 1$\times1$ convolution before the softmax layer. All the other convolutional layers are composed of 3$\times$3 filters. The dilation rate $D$ is shown at the bottom of the convolutional layers in the first block. In the rest of the network, blocks with the same color correspond to the same dilation rate.

\section{Experiments}
\label{sec:Experiments}

\subsection{Materials}

\subsubsection{Patients population}

The study was approved by the Ethics Committee of Tangdu Hospital of the Fourth Military Medical University. Informed content was obtained from each enrolled subject. Sixty patients with pathologically-confirmed BC lesions between October 2013 and May 2016 were involved in this study (Table \ref{table:Patient cohort}). Among them, 12 patients had multiple focal bladder tumors, including 9 patients with two tumor sites, two patients with three tumor sites, and one patient with four tumor sites. A total of 76 BC lesions were identified, their diameter in the range of 0.57 - 6.05 cm (Table \ref{table:Patient cohort}).

\begin{table}[t!]
\centering
\footnotesize
\caption{Clinico-pathological characteristics of BC patients employed in the current study.}
\label{table:Patient cohort}
\setlength{\tabcolsep}{10pt}
\renewcommand{\arraystretch}{1.1}
\begin{tabular}{lc}
\toprule
\textbf{Characteristic} & \textbf{Value} \\
\midrule \midrule
Patients, No.(\%) &  60 (100\%) \\
Male, No.(\%) & 53 (88.33\%)\\
Female, No.(\%) & 7 (11.67\%)\\
Age, Median (Range), yrs &  67 [42, 81]\\
Tumors, No.(\%) &  76 (100\%) \\
Stage T1 or lower,  No.(\%) & 16 (21.05\%)\\
Stage T2,           No.(\%) & 42 (55.26\%)\\
Stage T3 or higher, No.(\%) & 18 (23.69\%)\\
Tumor Size, Median (Range), cm  &   2.22 [0.57, 6.05]\\
\midrule
\end{tabular}
\end{table}

\subsubsection{Image acquisition}

All subjects were examined before treatment by a clinical whole body MR scanner (GE Discovery MR 750 3.0T) with a phased-array body coil. A high-resolution 3D Axial Cube T2-weighted (T2W) MR sequence was adopted due to its high soft tissue contrast and relatively fast image acquisition. Prior to scanning, each patient was asked to drink enough mineral water and then waited for an adequate time period so that the bladder was sufficiently distended. The acquisition time ranged from 160.456 to 165.135 s with three-dimensional scanning. The repetition and echo time were 2500 ms and 135 ms, respectively. The imaging process contained from 80 to 124 slices per scan, each of size 512 $\times$ 512 pixels, with a pixel resolution of 0.5 $\times$ 0.5 mm$^2$. Moreover, the slice thickness was 1 mm, and the space between slices also 1 mm.

\subsubsection{Ground truth}

For each dataset, the urine, bladder walls and tumor regions were manually delineated by two experts with 9 years of experience in MR image interpretation, using a custom-developed package of MATLAB 2016b. Particularly, during the delineation process, all the target regions, including IW, OW and tumor regions, were first independently outlined slice-by-slice by the two experts who were blinded to the pathological results of the patient. Afterwards, both their delineations were mapped to the corresponding images with different contour colors. Then, the two experts worked together to achieve a consensus by modifying their delineations, referring to the corresponding pathological results or the functional MRI images (if available) with the corresponding slice identified by the registration of functional MRI with T2W image data sets. If disagreements on the delineations remained, the average of the two delineations was computed.

\subsubsection{Evaluation}

Similarity between two segmentations can be assessed by employing several comparison metrics. Since each of these yields different information, their choice is very important and must be considered in the appropriate context. The Dice similarity coefficient (DSC) \cite{dice1945measures} has been widely used to compare volumes based on their overlap. The DSC for two volumes $A$ and $B$ can be defined as 
\begin{equation}
	\mathrm{DSC} \ = \ \dfrac{2\left|A \cap B \right|}{|A|+|B|}
\end{equation}

However, volume-based metrics generally lack sensitivity to segmentation outline, and segmentations showing a high degree of spatial overlap might present clinically-relevant differences between their contours. This is particularly important in medical applications, such as radiation treatment planning, where contours serve as critical input to compute the delivered dose or to estimate prognostic factors. An additional analysis of the segmentation outline's fidelity is highly recommended since any under-inclusion of the target region might lead to a higher radiation exposure in healthy tissues, or vice-versa, an over-inclusion might lead to tumor regions not being sufficiently irradiated. Thus, distance-based metrics like the average symmetric surface distance (ASSD) were also considered in our evaluation. The ASSD between contours $A$ and $B$ are defined as follows:
\begin{equation}
\mathrm{ASSD} \ = \ \dfrac{1}{|A| + |B|} \, \left(\sum_{a \in A} \min_{b \in B} \, d(a,b) + \sum_{b \in B} \min_{a \in A} \, d(b,a) \right),
\end{equation}
where $d(a,b)$ is the distance between point $a$ and $b$.

\subsection{Implementation details}

All networks were trained from scratch by employing the Adam optimizer with standard $\beta$ values equal to 0.9 and 0.99 and minimizing the cross-entropy between the predicted probability distributions and the ground truth. Weights were initialized as in \cite{glorot2010understanding}. The learning rate was initially set to 1e$-4$, and then decreased by half each time encountering 20 epochs without improvement on the validation set. During training, four images were employed for each mini-batch. All three models were implemented in pyTorch \cite{paszke2017automatic} and experiments were ran on a machine equipped with a NVIDIA TITAN X with 12GBs of memory.

\subsection{Results}

\begin{figure*}[ht!]
     \begin{center}
   \includegraphics[width=1\linewidth]{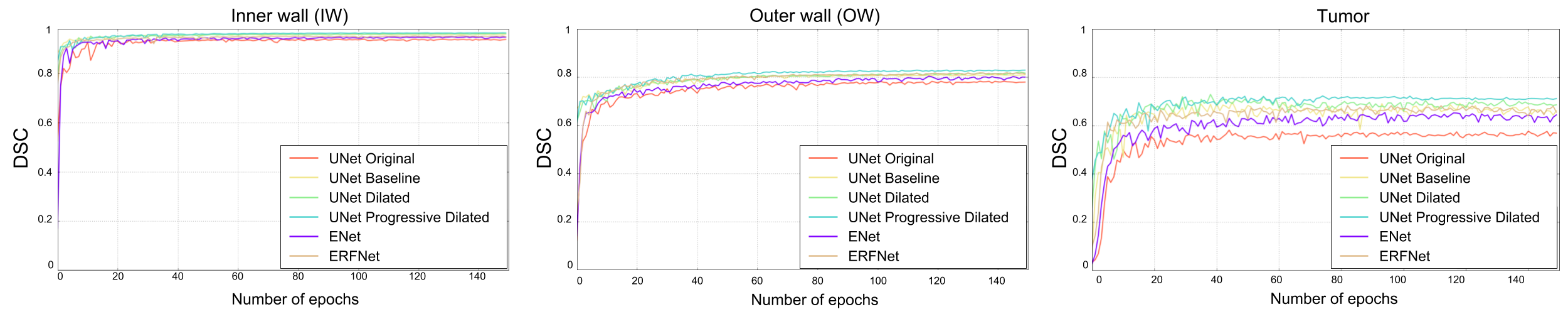}
\caption{Evolution of DSC on validation set during training for the inner and outer wall, as well as for tumor.} 
\label{fig:metrics_validation}
\end{center}
\end{figure*}

The performance of the UNet-Progressive model was compared with that of the UNet-Base and UNet-Dilated models introduced in Sections \ref{sssec:Unet} and \ref{sssec:Unet_Dilated}, as well as against the original UNet implementation \cite{ronneberger2015u}, ENet and ERFNet. For these experiments, the dataset was split into training, validation and testing sets composed of 40, 5 and 15 patients, respectively. These datasets remained the same for training and testing all models. Figure \ref{fig:metrics_validation} depicts the evolution of the DSC measured on the validation set at different training epochs. In general, all models yield similar performance for the inner wall (IW) and outer wall (OW) regions. However, in the case of tumor regions, the UNet-Progressive model obtained a higher accuracy than other models once the training converged.

Table \ref{table:metrics} reports the accuracy in terms of DSC and ASSD obtained by all evaluated models on the testing set. Results show the three models to achieve comparable results on inner and outer wall segmentation. However, as observed in the validation set, the UNet-Progressive model performed better than the baseline and UNet-Dilated when segmenting the tumor. Results from a one-tailed Wilcoxon signed-rank test between the proposed network with progressive dilated convolutions and other models are shown in Table \ref{table:mystatistics}. We see that our UNet-Progressive network is statistically superior to standard UNet, for all regions and metrics. In all but one case, it also gives statistically better performance than ENet, which employs similar dilated modules. The advantage of our method is particularly substantial for tumor regions, where it statistically outperforms all compared methods except UNet-Dilated. Improvements for this region are especially pronounced for the ASSD metric (see Table \ref{table:metrics}), which accounts for clinically-relevant differences between contours. 

\begin{table*}[t!]
\centering
\footnotesize
\caption{Mean DSC and ASSD values obtained by evaluated methods on the independent testing group.}
\label{table:metrics}
\setlength{\tabcolsep}{4pt}
\renewcommand{\arraystretch}{1.1}
\begin{tabular}{lccccccc}
\toprule
  & \multicolumn{3}{c}{\textbf{DSC}}   & \multicolumn{1}{l}{} & 
  \multicolumn{3}{c}{\textbf{ASSD (mm)}}    \\ \cmidrule(l{5pt}r{5pt}){2-4} \cmidrule(l{5pt}r{5pt}){6-8}
 & Inner Wall  & Outer Wall  & Tumor   &    & Inner Wall  & Outer Wall  & Tumor     \\
 \midrule \midrule
UNet-Original & 0.9701 $\pm$ 0.0130 & 0.7969 $\pm$ 0.0492 & 0.5638 $\pm$ 0.1646 & &  0.4260 $\pm$ 0.1749 & 0.5590 $\pm$ 0.1673 & 4.4298 $\pm$ 2.6382 \\
UNet-Baseline    &0.9839 $\pm$ 0.0030 & 0.8344 $\pm$ 0.0214 & 0.6276 $\pm$ 0.0963 &   & 0.3379 $\pm$ 0.0796 & 0.4503 $\pm$ 0.0919& 3.7432 $\pm$ 1.6923 \\
UNet-Dilated     & \textbf{0.9844} $\pm$ 0.0030 & 0.8386 $\pm$ 0.0232& 0.6791 $\pm$ 0.0818 & &  \textbf{0.3210} $\pm$ 0.0632&  \textbf{0.4238} $\pm$ 0.0725& 3.4320 $\pm$  1.9224\\
UNet-Progressive* &   0.9836 $\pm$ 0.0033 & \textbf{0.8391} $\pm$ 0.0247& \textbf{0.6856} $\pm$ 0.0827  &    &    0.3517 $\pm$ 0.0874& 0.4299 $\pm$ 0.0859& \textbf{2.8352} $\pm$ 1.1865 \\
\midrule
ENet \cite{paszke2016enet} & 0.9788 $\pm$ 0.0082 & 0.8065 $\pm$ 0.0446 & 0.6185 $\pm$ 0.1436 &      &  0.3775 $\pm$ 0.1245 & 0.5046 $\pm$ 0.1452 & 4.3640 $\pm$ 3.5614  \\
ERFNet \cite{romera2017efficient} & 0.9822 $\pm$ 0.0038 & 0.8367 $\pm$ 0.0378 & 0.6412 $\pm$ 0.1192  &  &  0.3398 $\pm$ 0.1069 & 0.4377 $\pm$ 0.1314 & 3.4588 $\pm$ 2.1505 \\                
\bottomrule\\[-4pt] 
\multicolumn{7}{l}{* UNet-Progressive corresponds to our proposed method.}
\end{tabular}
\end{table*}

\begin{table*}[t!]
\centering
\footnotesize
\caption{Wilcoxon signed-rank test between the proposed network with progressive dilated convolutions and other investigated networks. Significant differences (p\,$<$\,0.05) are highlighted in bold.}
\label{table:mystatistics}
\setlength{\tabcolsep}{10pt}
\renewcommand{\arraystretch}{1.1}
\begin{tabular}{lcccccc}
\toprule
\textbf{Region}  & \textbf{Metric} & \textbf{UNet-Original} & \textbf{UNet-Baseline} & 
	\textbf{UNet-Dilated} & \textbf{ENet}  & \textbf{ERFNet} \\ 
\midrule\midrule
\multirow{2}{*}{Inner Wall} & Dice & \textbf{0.001}         & 0.865         & 0.256        & \textbf{0.005} & 0.156  \\
 & ASSD & \textbf{0.020}         & 0.691         & 0.112        & 0.394 & 0.496  \\
\midrule
\multirow{2}{*}{Outer Wall} & Dice & \textbf{0.001}         & 0.156         & 0.691        & \textbf{0.001} & 0.496  \\
 & ASSD & \textbf{0.006}         & 0.173         & 0.570        & \textbf{0.012} & 0.609  \\
\midrule
\multirow{2}{*}{Tumor} & Dice      & \textbf{0.002}         & \textbf{0.005}         & 0.910         & \textbf{0.012} & \textbf{0.031}  \\
 & ASSD      & \textbf{0.001}         & \textbf{0.006}         & 0.460        & \textbf{0.012} & \textbf{0.047}  \\ 
\bottomrule
\end{tabular}
\end{table*}

\begin{figure*}[ht!]
     \begin{center}
     \mbox{
        \includegraphics[width=1\linewidth]{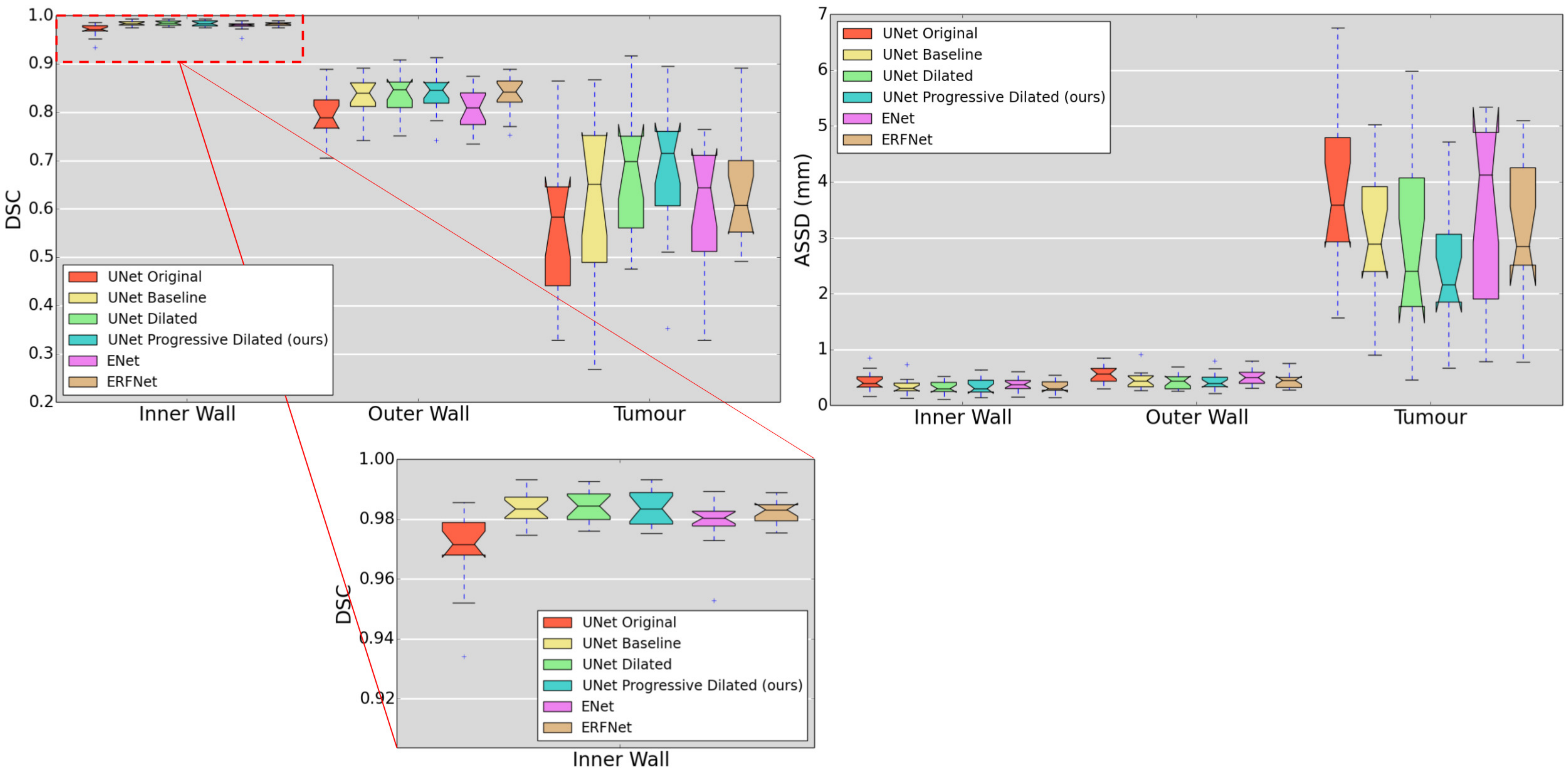}  
        }
        \caption{Distribution of DSC and ASSD values obtained by tested networks.}
\label{fig:metrics}
\end{center}        
\end{figure*}

Figure \ref{fig:metrics} details the distribution of DSC and ASSD values for tested models. In these plots, we can first observe how adding dilated convolutions to standard UNet architectures improves performance, which is reflected in the distribution of segmentation accuracy values. Further, if convolutions are added progressively, such as in the proposed modules, the distribution of values remains more compact (i.e., the variance is smaller). This is more prominent for results on tumor regions in both DSC and ASSD distribution plots.

\begin{figure*}[ht!]
\centering
\includegraphics[width=.95\linewidth]{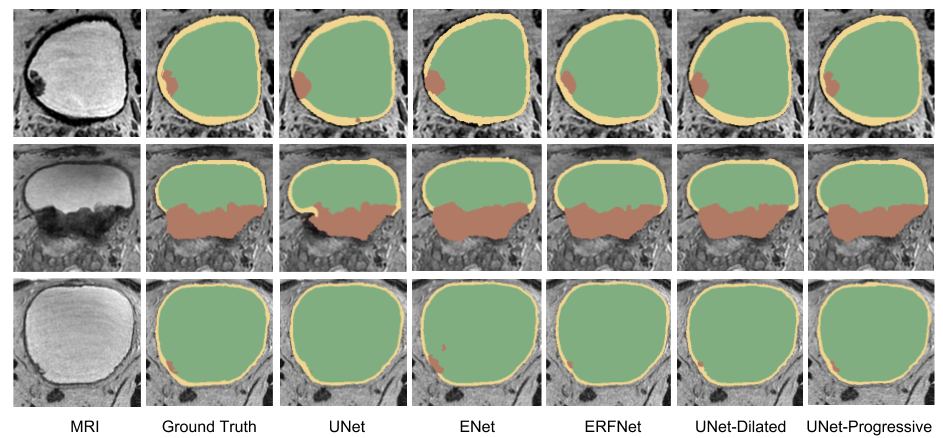}
\caption{Visual results achieved by all the models. While inner and outer walls are respectively represented by green and yellow regions, tumour regions are highlighted in brown.}
\label{fig:bladderRes}
\end{figure*}

To visualize the impact of including dilated convolutions in the standard way, or including progressive dilated modules, Fig. \ref{fig:bladderRes} shows segmentation results of the baseline UNet, UNet with standard dilated convolutions, and the proposed model. These results illustrate the variable sizes of tumors, some of them quite small and thus hard to segment (e.g., the tumor in the bottom row). Once again, we see that the three models achieve similar segmentations for inner and outer walls, and that differences arise when comparing the tumor segmentations provided by the models. Even though the tumor is typically identified by all the models, the proposed UNet-Progressive model achieves the most reliable contours compared to the ground truth. UNet underestimates the tumor region in two of the three examples, and generates a blobby contour in the third case (\emph{top}). On the other hand, UNet-Dilated improves results compared to the version without dilated convolutions, however fails to separate outer walls from carcinogenic regions in some cases (\emph{top of the figure}). By employing progressive dilated modules, our UNet-Progressive network can successfully differentiate tumor and outer walls, as shown in the \emph{top-right} image of Fig. \ref{fig:bladderRes}.

To show that adding the proposed progressive dilated convolution modules does not introduce a burden on computation time, we compared the different UNet-based architectures in terms of efficiency (Table \ref{table:times}). We observe that inference times per 2D slice is very similar across the three deep models. Taking into account that a volume contains between 80 and 124 2D slices, the segmentation of a whole volume was performed in less than a second, regardless the architecture.

\begin{table}[t!]
\centering
\caption{Inference times for the analyzed architectures.}
\label{table:times}
\setlength{\tabcolsep}{10pt}
\renewcommand{\arraystretch}{1.1}
\begin{tabular}{lc}
\toprule
\textbf{Method}  & \textbf{Mean time (ms\,/\,slice)} \\ 
\midrule\midrule
UNet-Original &  4.11 $\pm$ 0.61   \\
UNet-Baseline & 4.42 $\pm$ 0.64   \\
UNet-Dilated & 5.22 $\pm$ 1.04  \\
UNet-Progressive & 5.87 $\pm$ 1.13 \\ 
ENet \cite{paszke2016enet} &  3.24 $\pm$ 0.52   \\
ERNet \cite{romera2017efficient} & 3.83 $\pm$ 0.54   \\
\bottomrule
\end{tabular}
\end{table}

\section{Discussion}

In this study, a deep CNN model with progressive dilated convolutional modules was proposed to segment multiple regions in MRI images of BC patients. The proposed network extends the well-known UNet model by including dilated convolutions where the dilation rate within each module increases progressively. We evaluated our model on MRI image datasets acquired from an in-house cohort of 60 patients with BC. Results demonstrate the proposed approach to achieve state-of-the-art accuracy compared to existing approaches for this task, in a fraction of time. Additionally, when compared against similar networks, our architecture also demonstrated outstanding performance, particularly for tumor regions.

Tested models have shown similar results for the segmentation of the inner and outer bladder walls. However, a significant improvement was observed between the UNet-based models for tumor segmentation, particularly between the baselines and models with dilated convolutions. This improvement is due to the larger receptive field provided by dilated convolutions, which leverages more contextual information. When using progressive dilated convolutions, the ability to span similar-sized regions while avoiding large dilation rates -- which insert many holes between neighbor pixels -- might explain improved accuracy compared to models with standard dilated convolutions.

The proposed model also outperformed recent deep architectures that incorporate dilated convolutions with different dilation rates, such as ENet or ERFNet. It is important to note that those networks were developed to achieve a trade-off between performance and inference time. Therefore, their limited number of learnable parameters may explain differences with the architecture proposed in this work. Nevertheless, unlike out model, these networks do not implement skip connections between layers from the encoder and the decoder, which was shown to be important for recovering spatial information lost during downsampling \cite{drozdzal2016importance}.

Direct comparison with previous state-of-the-art methods for this task is challenging. First, as shown in Table \ref{table:methods}, research on automatic segmentation of multiple bladder regions in MRI images remains very limited. Moreover, different performance metrics were used to evaluate the performance of the few existing approaches, and some works only reported qualitative results \cite{garnier2011bladder,han2013unified}, which is subject to user interpretation. This makes it difficult to perform a fair and complete comparison between the proposed model and previous approaches for this problem. For example, Ma et al. \cite{ma2011novel} reported a mean DSC of 0.97 for the IW, but performance on the OW was not assessed. More recently, Qin et al. \cite{qin2014adaptive} evaluated their method on 11 subjects, reporting mean DSC values of 0.96 and 0.71 for IW and OW, and an average surface distance (ASD) of 1.45 and 1.94 for IW and OW, respectively. In another study, Xu et al. \cite{Xu2017} achieved a mean DSC of 0.87 measured on IW and OW. In light of the advantages of our models with respect to the state-of-art, we believe that approaches similar to those proposed in this work should now be considered to assess the segmentation of BC images. 

\begin{figure}[ht!]
\centering
\includegraphics[width=1\linewidth,trim={15mm 10mm 15mm 0},clip]{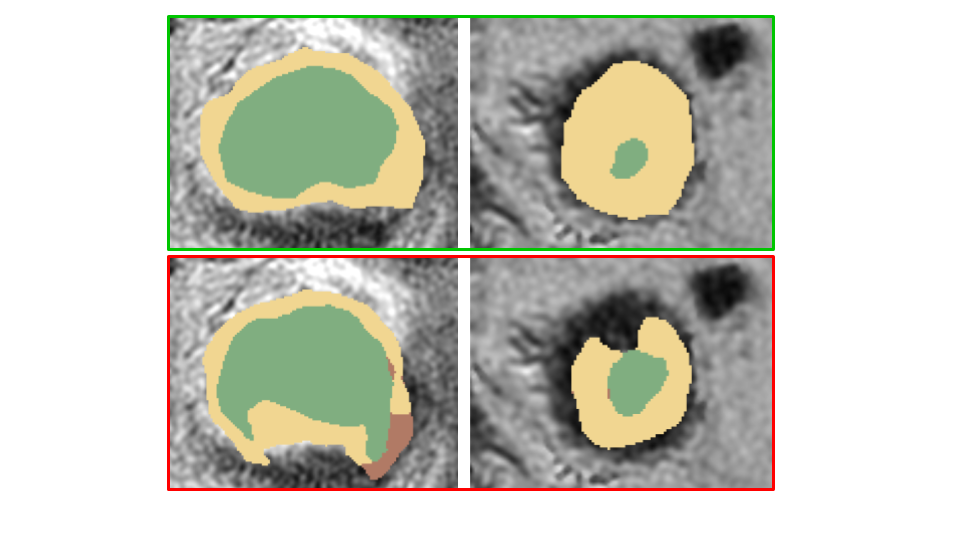}
\caption{Visual examples of segmentation failure. Ground truth are shown in the green frame (\emph{top}), whereas CNN segmentations are depicted in the red frame (\emph{bottom}). Green and yellow regions represent the inner and outer wall, respectively, and tumour regions are depicted in brown.}
\label{fig:fail}
\end{figure}

Although our results demonstrated the high performance of proposed models, there are some regions where segmentation might not be satisfactory in a clinical setting (e.g., Fig. \ref{fig:fail}). Particularly, these situations are observed in both the upper and lower extremes. We believe that these regions are more challenging to segment because of the thicker appearance of the walls. As segmentation is performed in 2D slices, there exist an imbalance in the number of samples representing the extremes of the urine --where outer walls are thicker in these images (See Fig \ref{fig:fail}, right column)-- with respect to middle regions. This might bias the CNN towards providing thinner wall regions across all the 2D slices. To improve CNN-based segmentation, recent works have considered to cast the probability maps from the CNN as unary potentials in an energy minimization framework \cite{Fechter_Esophagus, chen2018deeplab, kamnitsas2017efficient}. In these works, length is typically employed as a regularizer in the energy function. More complex regularizers have demonstrated to further boost the performance of segmentation techniques, e.g. convexity \cite{gorelick2017convexity} or compactness \cite{dolz2017unbiased}. Employing such regularizers may improve performance in the current application given the compact shape of the bladder. Furthermore, recent works have shown that combining several deep models can lead to important improvements in several segmentation tasks \cite{kamnitsas2017ensembles,dolz2017deep,manjon2018mri}. This promising strategy could also be investigated to improve performance, especially for the difficult task of segmenting bladder tumor regions. 

Given the 3D nature of the volumetric data in this particular application, by performing 2D segmentation in a slice-manner the anatomic context in directions orthogonal to the 2D plane is discarded. Adopting a 3D approach is, a priori, more appropriate in order to account for volumetric information. Nevertheless, in the context of the proposed approach, the large receptive field achieved by the proposed network makes the dimension of the input patch in 3D intractable. Due to memory limitations in current resources, 3D convolutional neural networks typically take input sub-volumes of size 64$\times$64$\times$64. Taking into account that the proposed progressive dilated network enlarges the receptive field, the input patch should be, at least, equal to 267 pixels per dimension, which cannot be efficiently allocated in our GPU facilities.

A main limitation of the current study is the limited available dataset employed in our experiments. First, the reduced amount of subjects employed for training (i.e., 45 MRI scans) is insufficient to capture the high variability of tumor regions across the population, as demonstrated by the results. Secondly, the data was acquired by the same scanner with the same imaging parameters, which may possibly reduce the generalization of the proposed scheme and impair its overall performance in segmentation. A larger validation group, including datasets acquired from multiple clinical centers with different scanners and imaging parameters, would further demonstrate its potential in real clinical applications. 

Even though segmentation is a fundamental task in the medical field, it rarely represents the final objective of the clinical pipeline. In the assessment of bladder cancer patients, segmentation of IW, OW and tumor is employed to evaluate the muscle invasiveness and grade of BC, which play a crucial role in treatment decision and prognosis \cite{Xu2017Preoperative,Zhang2017Radiomics,Liu2017Relationship,Xu2017}. In future works, we aim to apply the proposed multi-region segmentation scheme with a radiomics strategy for the pre-operative and automatic evaluation of BC. 
 
\section{Conclusion}

We proposed an approach using progressive dilated convolution blocks for the multi-region semantic segmentation of bladder cancer in MRI. Progressive dilated blocks allow having the same receptive field as standard dilated blocks but with lower dilation rates. The proposed network achieved a higher accuracy than approaches using standard dilated convolutions, particularly when segmenting tumors. Moreover, the proposed model outperformed state-of-the-art methods for the task at hand, bringing three important advantages: i) it enables segmenting multiple BC regions simultaneously, ii) there is no need for contour initialization, and iii) it is computationally efficient, e.g., 2-3 orders of magnitude faster than current approaches based on level-sets. In summary, deep CNNs in general, and the proposed network in particular, are very suitable for this task.
  
\ifCLASSOPTIONcompsoc
 \section*{Acknowledgments}
\else
 \section*{Acknowledgment}
\fi

This work is supported by the National Science and Engineering Research Council of Canada (NSERC), discovery grant program, the National Nature Science Foundation of China under grant No.81230035, National Key Research and Development Program of China under grant No.2017YFC0107400, Key project supported by Military Science and Technology Foundation under grant No.BWS14C030, and by the ETS Research Chair on Artificial Intelligence in Medical Imaging.

\ifCLASSOPTIONcaptionsoff
 \newpage
\fi

\bibliographystyle{IEEEtran}
\bibliography{refs,MRCMF_bladder,Xu_articles,New_added,Lambin2017}

\end{document}